%% file: main.tex
\pdfoutput=1

\documentclass[11pt]{article}
\usepackage[]{authblk}
\usepackage[]{ACL2023}

\usepackage{times}
\usepackage{latexsym}

\usepackage[T1]{fontenc}

\usepackage[utf8]{inputenc}

\usepackage{microtype}

\usepackage{inconsolata}

\usepackage{amsmath,amssymb,epsfig,marvosym}
\usepackage{graphicx}
\usepackage{capt-of}
\usepackage{booktabs}
\usepackage{varwidth}
\usepackage{subcaption}
\usepackage{arydshln}
\usepackage{bbm}

\usepackage{algorithm}
\usepackage[noend]{algpseudocode}

\title{Consistent Text Categorization using Data
Augmentation in e-Commerce}

\author[1,\thanks{\ \ The work was carried out during an internship at Yahoo Research.}]{Guy Horowitz}
\author[2]{Stav Yanovsky Daye}
\author[2]{Noa Avigdor-Elgrabli}
\author[2]{Ariel Raviv}

\affil[1]{Technion -- Israel Institute of Technology}
\affil[2]{Yahoo Research}
\affil[ ]{\texttt{guy.h@campus.technion.ac.il}}
\affil[ ]{\{\texttt{stav.yanovsky,noaa,arielr\}@yahooinc.com}}

\input{macros}

\begin{document}
\maketitle

\input{Chapters/1.abstract.tex}
\input{Chapters/2.introduction.tex}

\input{Chapters/4.problem_setup.tex}

\input{Chapters/5.method.tex}
\input{Chapters/7.experiments.tex}

\input{Chapters/8.results.tex}

\input{Chapters/9.conclusions.tex}

\section*{Limitations}

Our work has several limitations. First, our consistency study focuses on our used categorization model and was conducted on only one specific dataset. It might not perfectly generalize to other problems. Second, the proposed solutions are based solely on data augmentation without changing the current production settings and model. Other approaches such as changing the model's objective function to take consistency into account might also benefit the solution. Lastly, in terms of user perspective, while our solution show significant improvement over the baseline, inconsistencies are still visible.



\section*{Ethics Statement}

This NLP research study was designed and carried out with strict adherence to ethical principles and guidelines. The study was reviewed and approved by our company's research lead prior to the submission.
The study followed the ACL conference’s guidelines on the use of language data. The researchers take full responsibility for ensuring the ethical conduct of this study and are committed to upholding the highest standards of ethical research practices in NLP.


\bibliography{anthology,bibliography}
\bibliographystyle{acl_natbib}

\newpage
\appendix
\input{Chapters/appendix.tex}

\end{document}

%% file: macros.tex
\newcommand{\orange}[1]{{\leavevmode\color[RGB]{229,83,0}{#1}}}

\newcommand{\ariel}[1]{\orange{}} 

\newcommand{\specialcell}[2][c]{%
  \begin{tabular}[#1]{@{}c@{}}#2\end{tabular}}

\newcommand\eqexp[2]{\underset{#1}{\mathbbm{E}}{\left[ {#2} \right]}}
\newcommand\ind[1]{\mathbbm{1}\left\{#1\right\}}

\newcommand{\fbase}{{f^{\text{base}}}}
\newcommand{\Daug}{{\mathcal{D}_{\text{aug}}}}
\newcommand{\Dpairs}{{\mathcal{D}_{\text{pairs}}}}

%% file: Chapters/1.abstract.tex
\begin{abstract}

The categorization of massive e-Commerce data is a crucial, well-studied task, which is prevalent in industrial settings. In this work, we aim to improve an existing product categorization model that is already in use by a major web company, serving multiple applications.
At its core, the product categorization model is a text classification model that takes a product title as an input and outputs the most suitable category out of thousands of available candidates. Upon a closer inspection, we found inconsistencies in the labeling of similar items. For example, minor modifications of the product title pertaining to colors or measurements majorly impacted the model's output. This phenomenon can negatively affect downstream recommendation or search applications, leading to a sub-optimal user experience.

To address this issue, we propose a new framework for consistent text categorization. Our goal is to improve the model's consistency while maintaining its production-level performance. We use a semi-supervised approach for data augmentation and presents two different methods for utilizing unlabeled samples. One method relies directly on existing catalogs, while the other uses a generative model. We compare the pros and cons of each approach and present our experimental results.


\end{abstract}

%% file: Chapters/2.introduction.tex
\section{Introduction}


In the last two decades, widespread use of e-commerce platforms such as Amazon and eBay has contributed to a substantial growth in online retail. Such platforms rely on both explicit and implicit product features in order to deliver a satisfying user experience. There, the inferred product category is typically a crucial signal for many application such as browsing, search and recommendation. 

We focus on improving an existing product categorization model, we refer to as 'the categorizer', that is employed by our company for fast categorization of billions of items on a daily basis. It classifies e-commerce items, such as products or deals, based on a predefined hierarchy of categories, namely GPT (Google Product Taxonomy). Given a product title, the categorizer assigns the most appropriate label in the taxonomy. The model itself is highly scalable and effective, so it is well-suited for settings with large and rapidly growing item catalogs. In our company, the categorizer is used as a standalone component in various e-commerce related services, such as recommendation, search, and ad ranking.


A recent examination of the categorizer's output revealed inconsistencies in the labeling of similar items. It was evident that in some cases small variations in product titles, such as those relating to colors or measurements, significantly affect the categorizer's output. This inconsistency negatively impacts search and recommendation algorithms that rely on the inferred category, leading to a poor user experience.

The concept of consistency in NLP tasks has been studied in various research works, including robustness to paraphrasing~\cite{elazar2021measuring} and robustness to adversarial attacks~\cite{jin2020bert,wang2020cat}. 
Other works relate consistency issues with the misuse of  spurious features during the learning phase~\cite{arjovsky2019invariant, veitch2021counterfactual,wang2021identifying}. 

When examining the performance of the categorizer in terms of accuracy alone, the inconsistency issue may be overlooked. But, since many recommendation pipelines depend on the output of the product categorizer, an inconsistent model can have severe implications on the user experience. 
In most cases, the differences include returning the parent category or a sibling category, rather than a completely different category path.

To tackle this inconsistency problem, we use different \textit{data augmentation} techniques and enrich the training data with item versioning, leading to a more consistent model. 
Data augmentation for improving various NLP tasks has been widely studied and surveyed \cite{shorten2021text}, and particularly in the context of consistency \cite{xie2020unsupervised}. Generating such data, both manually~\cite{kaushik2019learning} and automatically~\cite{rizos2019augment,bari2020uxla, kumar2020data}, has shown to contribute to the robustness of learnt models in different settings. We chose to use data augmentation, without changing the current architecture of the already-in-use product categorizer for two main reasons. First, for scalability reasons, any change in the architecture might degrade the model's ability to infer the categories of billions of items per day. Second, maintaining the current model architecture expedites the productization process and requires only minimal engineering effort.

This work defines a new framework, {\it Consistent Semi-Supervised Learning (Consistent-SSL)}, for consistent text categorization in the context of e-commerce (Section \ref{sec:consistent}). We use an unlabeled clustered dataset as a source of legit item versioning. The dataset is derived from product catalogs, and includes clusters of different versions of items. We present two different methods to utilize this unlabeled clustered data: a self-training method and a generative approach (Section \ref{sec:methods}). 
We describe the datasets and the experimental framework we use for the evaluation of the proposed methods (Section \ref{sec:eval}). Finally, we detail results, showing an improvement in the consistency rate of 4-10\% above the baseline model, and discuss the advantages and weaknesses of each method (Section \ref{sec:results}).\looseness=-1

%% file: Chapters/4.problem_setup.tex
\section{Consistent Semi-Supervised Learning}\label{sec:consistent}
We now formalize our notion of consistent classification and introduce the settings for consistent Semi-Supervised Learning (consistent-SSL).

\subsection{Consistent Classification}
In order to formalize consistent classification, let $ \mathcal{X}$ be our set of items, and  $\mathcal{Y}= [c]$ for $c\in \mathbb{N}$, be a final set of labels.  Each item $ x\in \mathcal{X}$ corresponds to a label $y \in \mathcal{Y}$. 

Additionally, let  $\mathcal{V}: \mathcal{X} \rightarrow \mathcal{X}$, be a non-deterministic perturbation function which transforms an item from one version $x$ to another $\hat{x}$. For example, if $x= \text{"blue T-shirt small size"}$, $\hat{x}\sim \mathcal{V}(x)$ could be $\hat{x}=\text{"black T-shirt small size"}$ or $\hat{x}=$ "blue T-shirt large size". 
We assume that the perturbation function is label-preserving, i.e. $x,\hat{x}\sim \mathcal{V}(x)$ share the same label $y$.
Let $p(x,y)$ be a joint distribution over items and labels and $p(x)$ the marginal distribution over items.
The goal of consistent classification is to learn a classifier $f : \mathcal{X}  \rightarrow \mathcal{Y}$ from a class $F$ with a dual objective: a high expected \textbf{accuracy}, i.e. high expected value of the indicator that an item $x\in X$ is labeled by $f$ to its correct label $y$: 
\begin{equation}
\label{eq:accuracy}
\eqexp{(x, y)\sim p(x,y)}{\ind{f(x) = y}}
\end{equation}
 and a high expected \textbf{consistency}, which we define as:
\begin{equation}
\label{eq:consistency}
\eqexp{\substack{x\sim p(x), \\ \hat{x}\sim \mathcal{V}(x)}}{\ind{f(x) = f(\hat{x})}}
\end{equation}
i.e. the expected value of the indicator of two items $x,\hat{x}\sim \mathcal{V}(x)$ to be transformed by $f$ to the same label.
Therefore, the dual objective of $f$ can be formalized as:
\begin{equation}
\label{eq:dual objective}
\begin{aligned}
\min_{f} \eqexp{\substack{(x, y)\sim p(x,y), \\ \hat{x}\sim \mathcal{V}(x)}}{\ind{f(x) \neq y} + \lambda \ind{f(x) \neq f(\hat{x})}}    
\end{aligned}
\end{equation}
where $\lambda \in \mathbb{R}$ controlling the balance between the accuracy loss and the consistency loss.

Note that there could be a trad-off between the accuracy objective and the consistency one. A model that is trained to disregard specific features like color or size would be more consistent but as those features might be informative in order to partition between some categories this could harm the overall accuracy. For example, if the color purple is more likely to appear in sport shoes than in evening shoes, a model that is trained to give less weight to colors may have a harder time distinguishing between sports and evening shoes while being more robust to changes in colors and thus more consistent.

\subsection{ Consistent-SSL Settings}
In SSL settings, we are given labeled data $\mathcal{D}_{L}=\{(x_i, y_i)\}_{i=1}^{l}$, which is assumed to be sampled i.i.d. from $p$, and unlabeled data $\mathcal{D}_{U}=\{x_i\}_{i=l+1}^{l+u}$ possibly sampled from another distribution $q$. 
We tune a classifier $f$ using both $\mathcal{D}_L$ and $\mathcal{D}_U$. 

This work extends the standard SSL settings to consistent-SSL.
The unlabeled data $\mathcal{D}_U$ is clustered with respect to the perturbation function $\mathcal{V}$, i.e. it consists of $u$ sets of items $X_i$, each set contains $k_i$ versions $\hat{x}_{j}^{(i)} \sim \mathcal{V}(x_i)$ of the same item $x_i$.
More formally, $\mathcal{D}_U = \left\{X_i\right\}_{i=l+1}^{l+u}$, where, $X_i=\left\{\hat{x}_j^{(i)} \right\}_{j=1}^{k_i}$, and $\hat{x}_j^{(i)}\sim \mathcal{V}(x_i)$ for $j=1\ldots k_i$.

The goal in consistent-SSL is to learn a classifier $f$ that optimizes the objective in Eq. \eqref{eq:dual objective} given $\mathcal{D}_L$ and $\mathcal{D}_U$. 
Note that $\mathcal{V}$ is unknown, and only appears indirectly in the $\mathcal{D}_U$ samples.

%% file: Chapters/5.method.tex
\section{Methods}\label{sec:methods}
We present two methods for consistent-SSL, Consistent Self Training (CST) and Consistent Generative Augmentation (CGA).
Both methods utilize the unlabeled samples from $\mathcal{D}_U$ for data augmentation. In each method we create an augmented set $\Daug$ using $\mathcal{D}_U$ and train a classifier $f$ on $\mathcal{D}_L\cup \Daug$.
This approach optimizes indirectly the objective of Eq.~\eqref{eq:dual objective}, as we add additional training samples $\Daug$ that consists of different versions of the same items. The goal is to expose $f$ to a more diverse set of item versions in training time, making it more robust to minor changes.

Let us review our approach using an illustrative example. Consider a dataset that contains clothing items. Assuming that $\mathcal{D}_L$, which was sampled from the distribution $p$, exhibits a spurious correlation between color of an item to its category (e.g. most of the black items are coats and most of the red items are dresses), then a classifier that was trained solely on $\mathcal{D}_L$ will tend to rely on the color of the item when it predicts its category.
When applying the model, $\mathcal{V}$ could change the items' colors and therefore the classifier will not be consistent (e.g. if $\mathcal{V}$ transforms a black coat to a red one, the classifier might predict different categories).
But, assuming the training data includes an item 
in multiple colors (e.g. black coat, red coat, blue coat, etc.), with the same label (e.g. \emph{Coats $\&$ Jackets}), then a model that is trained on such data will not relate a specific color to a specific label.
Such a model will be encouraged to ignore the color of an item when it predicts the label, and therefore will be more robust to changes in color. Note that colors here are only an example of one kind of versioning of items. Spurious features in the data could be related to colors, measurements, models, materials etc.

\subsection{Consistent Self Training (CST)}\label{subsec:cst}

\begin{figure}[h]
    \centering
    \vspace{-1em}
    \includegraphics[width=0.8\linewidth]{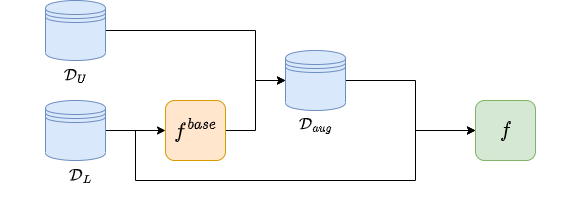}
    \vspace{-1em}
    \caption{
    Illustration of CST pipeline. A base model $\fbase$ is trained on the labeled training set $\mathcal{D}_{L}$. It is then used to assign pseudo labels for the unlabeled samples from $\mathcal{D}_{U}$ to create $\Daug$. A classifier $f$ is trained on $\mathcal{D}_{L}\cup \Daug$.
    }
\label{fig:CST_ill}
\end{figure}
In our first method, named Consistent Self Training (CST), we add samples from $\mathcal{D}_{U}$ to the labeled training data $\mathcal{D}_{L}$ and a new classifier $f$ is trained on the unified dataset.
Since the data of $\mathcal{D}_{U}$ is unlabeled, we perform a variant of self training \cite{lee2013pseudo, arazo2020pseudo, triguero2015self}.
To make sure that $\Daug$ is consistent, it's important that each item set $X_i$ is assigned with the same pseudo-label $\tilde{y}_i$.
To calculate $\tilde{y}_i$, we first train a base model $\fbase$ on the labeled data $\mathcal{D}_{L}$ and then use it to choose a single pseudo-label for each example set $X_i$, i.e. $\tilde{y}_i\gets h(X_i;\fbase)$, where $h$ is a function that given a set of examples and a classifier $\fbase$ returns a single label.  
For example, $h$ could return the prediction of $\fbase$ that got the highest confidence score, or the most frequent prediction across $X_i$. The function $h$ is an hyperparameter of the method. 
Finally, a classifier $f$ is trained over $\mathcal{D}_{L}\cup \Daug$.
Figure \ref{fig:CST_ill} shows an illustration of the CST pipeline, and a full description of the algorithm is presented in Appendix~\ref{alg:bootstrapping}.\looseness=-1

\subsection{Consistent Generative Augmentation (CGA)}\label{subsec:cga}

\begin{figure}[H]
    \centering
    \includegraphics[width=\linewidth]{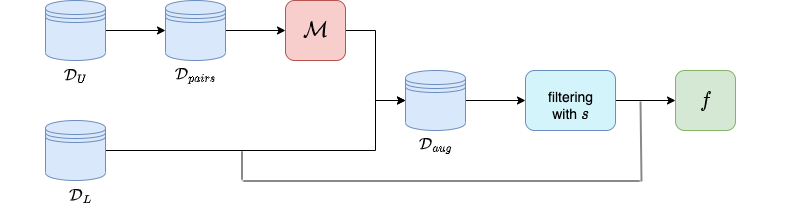}
    \caption{
    Illustration of CGA pipeline. A generative model $\mathcal{M}$ is trained on pairs of items from the catalog dataset $\mathcal{D}_{U}$. Then it is used to augment the labeled training set $\mathcal{D}_{L}$. The generated samples are filtered using a score function $s$. A classifier $f$ is trained on $\mathcal{D}_{L}\cup \Daug$.
    }
\label{fig:CGA_ill}
\end{figure}

We now detail our second method, we refer to as Consistent Generative Augmentation (CGA). Here, we train a generative model $\mathcal{M}$ on $\mathcal{D}_{U}$ in order to learn the perturbation function $\mathcal{V}$, and we use it to generate new samples based on the instances of $\mathcal{D}_{L}$.  
For this end, an item-pair dataset of different versions of items, $\Dpairs$ is constructed from $\mathcal{D}_{U}$; 
$\Dpairs = \left\{\left(\hat{x}_j^{(i)}, \hat{x}_{j'}^{(i)}\right) \\ \Big| l+1\leq i\leq l+u\bigwedge j,j'\in [k_i] \right\}.$
We train $\mathcal{M}$ on $\Dpairs$ to generate the second item given the first of each pair, while maintaining its label. Note that $\hat{x}_{j'}^{(i)} \sim \mathcal{V}(\hat{x}_j^{(i)})$.
Next, we generate an augmentation set $\Daug$ using $\mathcal{D}_{L}$ by applying $\mathcal{M}$ on each $(x, y)\in\mathcal{D}_{L}$ to get a new labeled sample $(\hat{x}, y)$.
Note that we can use $\mathcal{M}$ to generate multiple new samples from a single sample $x$. 
After creating $\Daug$, we filter it using a score function $s: \mathcal{X}\times \mathcal{X} \rightarrow [0,1]$ that aims to measure the quality of the generated $\hat{x}$ with respect to its origin $x$. Additionally, we remove low quality samples from $\Daug$ according to some predefined filter threshold $T$.
Finally, we train a classifier $f$ over $\mathcal{D}_{L}\cup \Daug$. 
Both $s$ and $T$ are hyper-parameter of the CGA method.
Figure \ref{fig:CGA_ill} shows an illustration of the CGA pipeline, and a full description of the algorithm is presented in Appendix~\ref{alg:gen}.
\subsection{ Methods Comparison}\label{subsec:comparison}

We compare the two proposed methods by three main aspects: the quality of the augmented product titles, the quality of the labels and the overall distribution. 

Considering the quality of the product titles, the CST method utilizes the unlabeled clustered data itself and thus provides product titles that are sampled from the real world and captures information about the true perturbation function $\mathcal{V}$. In contrast, the CGA method uses generated product titles, which may not represent $\mathcal{V}$ accurately. Regarding the label quality, the CGA method utilizes labels that are taken directly from the ground truth labels of the original items and thus of a better quality than the ones of the CST method, which uses calculated "pseudo-labels". With respect to the distribution of the data, the generated samples in the CGA method are taken directly from the distribution $p$ of the labeled training set. In contrast, in CST the unlabeled data comes from a distribution $q$ that is different than $p$, thus biasing the overall distribution of the training set.

The quality of the product titles in the augmentation set impacts the consistency 
and corollary the overall optimization of the model $f$. 
On the other hand, both the quality of the labels and the distribution of the augmentation set influence the accuracy 
which again affects the overall optimization of $f$.

%% file: Chapters/7.experiments.tex
\section{Empirical Evaluation}\label{sec:eval}

We now present our experimental results. We note that in all of our experiments, we use a model that is based on FastText~\cite{joulin2016bag} architecture, and has an hierarchical structure. This specific model is found to perform well on our task, as it takes into account the hierarchical structure nature of the labels. For more details, see Appendix~\ref{sec:HFT}.


\subsection{Train And Test Data}

\begin{figure*}[t]
  \centering
  \subcaptionbox{distribution of $\mathcal{D}_L$}{               \includegraphics[scale=0.25]{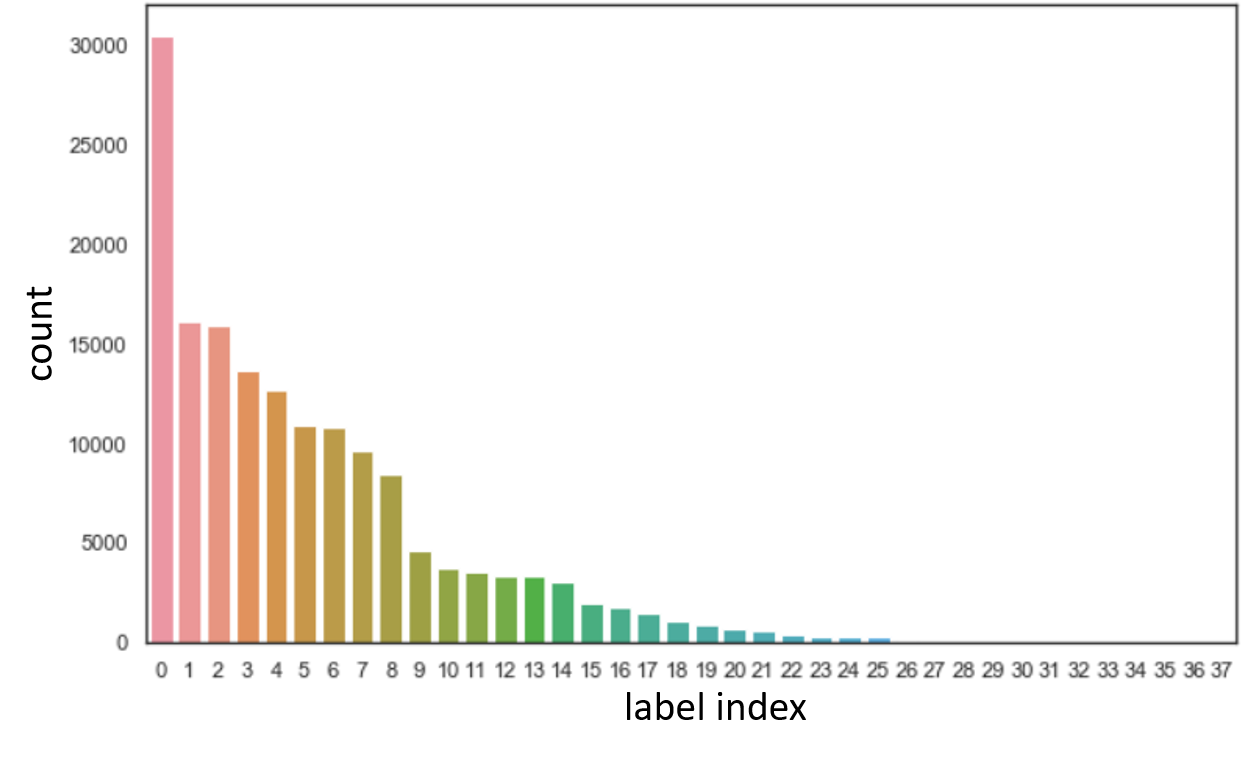}
  }
  \subcaptionbox{distribution of \textbf{complete} $\mathcal{D}_U$}{           \includegraphics[scale=0.28]{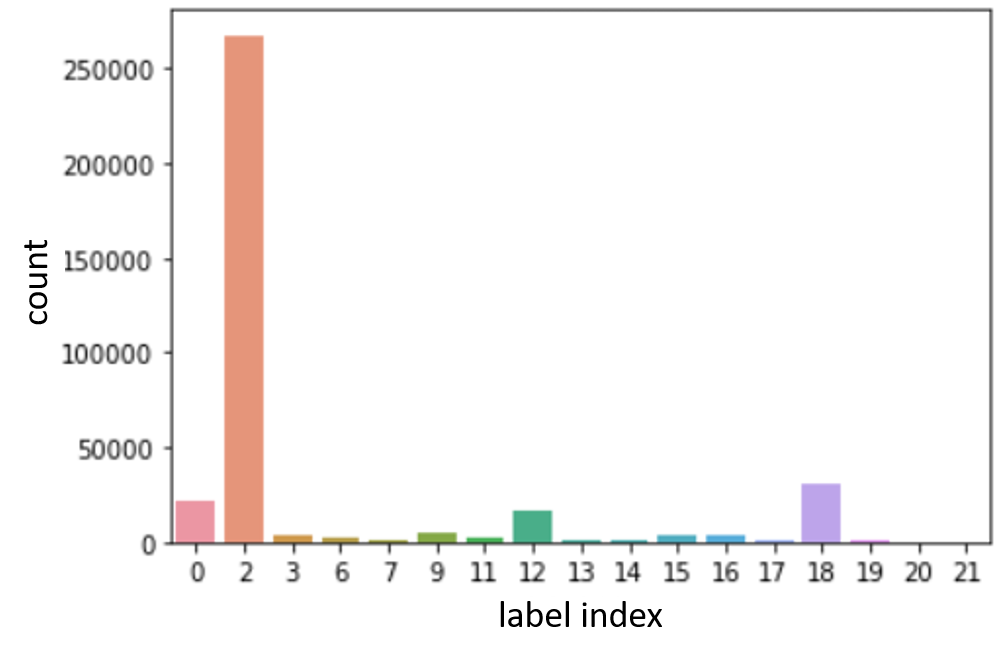}
  }
  \subcaptionbox{distribution of \textbf{sub sampled} $\mathcal{D}_U$}{              \includegraphics[scale=0.28]{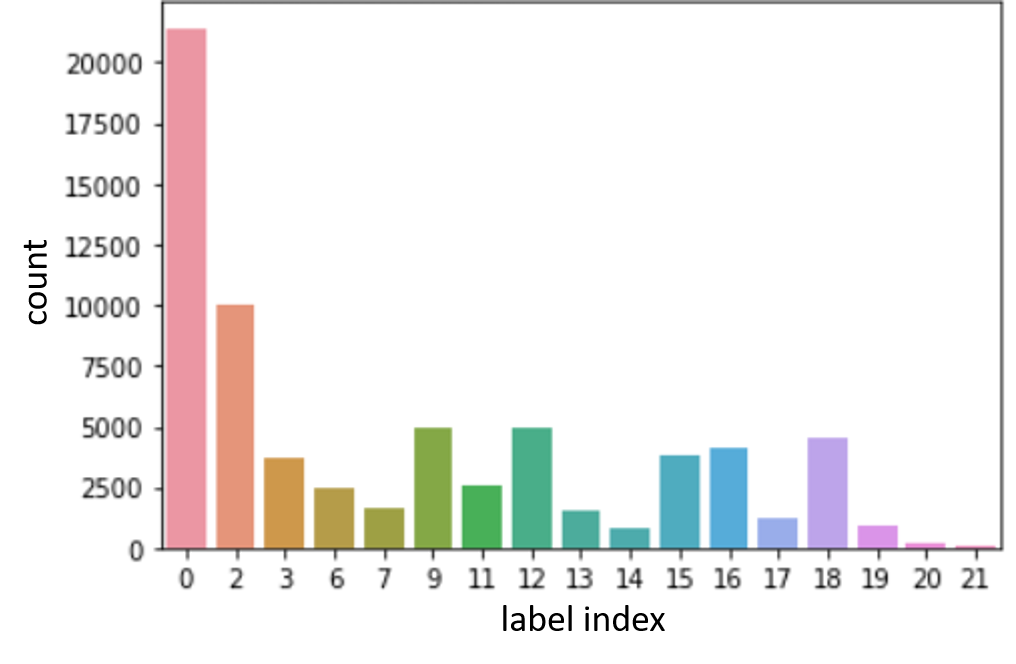}
  }
  \vspace{-0.5em}
  \caption{ 
  Distributions of the different versions of the data for CST. The labels are presented in $L_1$ granularity.} 
  \label{fig:distributions}
\end{figure*}

\begin{figure*}[t]
  \centering
  \subcaptionbox{different $T$, fixed $N$}{               \includegraphics[scale=0.3]{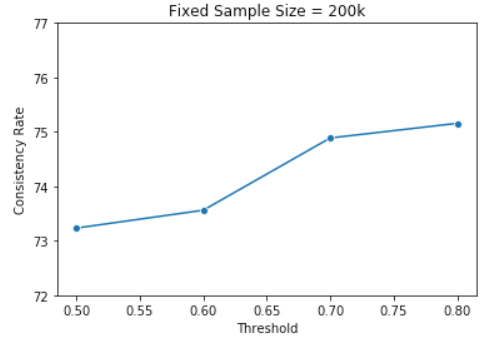}\label{fig:diffT}
  }
  \subcaptionbox{fixed $T$, different $N$}{               \includegraphics[scale=0.3]{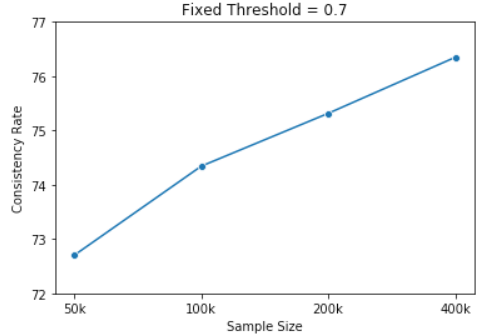}
  }
  \subcaptionbox{different $T$, different $N$}{           \includegraphics[scale=0.3]{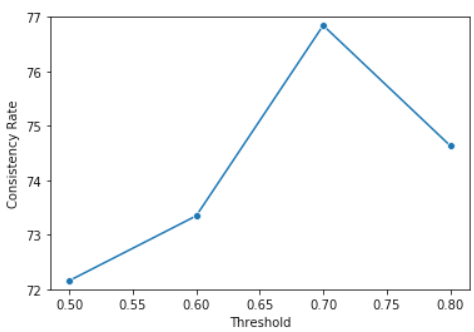}
  }
  \vspace{-0.5em}
  \caption{CGA experiments results.}
  \label{fig:t5_exp_graphs}
\end{figure*}

We conduct experiments using an e-commerce text classification dataset in order to empirically evaluate our methods. The items in this dataset are titles of commercial products, represented as free text, and the labels are the items' categories. The labels are taken from a hierarchic products taxonomy with 4 levels of granularity $\{L_i\}_{i=1}^{4}$. For example, consider a product title such as \emph{"Greenies Breath Buster Bites Fresh Flavor Grain-Free Dental Dog Treats, 1.2-oz bag"}, and its corresponding category \emph{Animals \& Pet Supplies $>$ Pet Supplies $>$ Dog Supplies $>$ Dog Treats}. 

Our dataset contains 184k labeled samples with 3k different labels, and additional 1.3M unlabeled samples.
The labeled samples correspond to real-world commerce related items, and are labeled by human annotators.
The unlabeled samples are retrieved from a product catalog of multiple retailers 
\ariel{rephrase}that includes grouping information. Each group contains multiple versions of the same item, e.g. \emph{"L.A. Girl, Matte Lipstick, \textbf{Snuggle}, 0.10 oz"} and \emph{"L.A. Girl, Matte Lipstick, \textbf{Bite Me}, 0.10 oz"}.
There are 363k different groups in the unlabeled catalog data, each group contains 2 to 192 items, and the average group size is 3.6. We note that the labeled and unlabeled data sets originate from different sources. This results in different category representation between the labeled and unlabeled data, e.g. several categories in the unlabeled data have low coverage compared to the labeled one. \looseness=-1

Our experiments measure both accuracy and consistency of the tested models.
To this end, we create two different test sets:

\textbf{Accuracy test.} The accuracy test is a standard test set that consists of labeled samples, on which we compute the weighted average F1 score of a given model.
The accuracy test contains 23k labeled examples sampled uniformly at random from the labeled data. We use the remaining 161k labeled samples as the $\mathcal{D}_L$.

\textbf{Consistency test.} The consistency test consist of pairs of item titles $(\hat{x}^1, \hat{x}^2)$, each pair includes two different versions of the same item. 
We define the \emph{consistency rate} of a given model $f$ to be the percentage of the $(\hat{x}^1, \hat{x}^2)$ pairs from the consistency test  that receive the same label prediction by $f$, i.e. $f(\hat{x}^1)=f({\hat{x}}^2)$.
We create this test set by sampling 9k  groups from the unlabeled data, then by sampling one pair of different titles $(\hat{x}^1, \hat{x}^2)$ from each group.
Since the consistency rate of a model on this test should be an empirical evaluation of its consistency as defined in Eq.~\eqref{eq:consistency}, the distribution of the data in this test should be similar to the distribution of the data in the accuracy test.
To mitigate some of the discrepancy between the unlabeled and labeled datasets, we sub-sample the unlabeled dataset according to the $L_1$ distribution of the labeled set. We use the unlabeled samples that are not selected for the consistency test as $\mathcal{D}_U$ for training.

\subsection{Experimental Framework}
This subsection describes in detail the configuration of the proposed methods, and the baselines that were used for comparison. 

\subsubsection{Baselines}
For the first \textbf{Baseline} model, we use the existing product categorization model, trained using only $\mathcal{D}_L$.
The second baseline is a \textbf{ColorsSizes-Blind (CS-Blind)} model. We train it using  $\mathcal{D}_L$ alone, while omitting colors and measurements from the data. We use predefined dictionaries of colors (e.g. "red", "white") and measurements (e.g. "small", "XL") to identify appearances in item titles and replace them with constant tokens, one for colors and another for sizes.
This baseline simulates an attempt to tackle the consistency issue by manually identifying few spurious features in the data and hiding them from the model to make it consistent. 

\begin{table*}[t!]
\centering
\begin{tabular}{|p{6.8cm}|p{7.cm}|c|}
\hline
Original Product Title & Generated Product Title & \specialcell{BLEU \\ score}\\ \hline
Polo Ralph Lauren Big Boys Fleece Hoodie & Polo Ralph Lauren Little Boys Fleece Hoodie & 0.795\\
Puff Sleeve T Shirt Ivory Frost & T Shirt & 0.135\\ 
Blackberries Prepacked 6 Oz & Cranberry Prepacked 6 Oz & 0.724\\
Sunnies Face Airblush in Peached & Sunnies Face Airblush in Peached Wall Poster With Pushpins & 0.482\\
Artistry Signature Color Long-wearing Eye Pencil Brown & Artistry Signature Color Long-wearing Eye Pencil Black & 0.850\\
\hline
\end{tabular}
\caption{Examples of pairs of original product titles and their corresponding generated ones, together with the computed BLEU score of the pairs.}
\label{table:generative_examples}
\end{table*}

\subsubsection{CST} \label{subsubsec: cst_empirical}

We evaluate CST with two configurations, each utilizes a different version of $\mathcal{D}_U$: 1) the \textbf{complete} data (354k groups with 1.3M samples), 
and 2) \textbf{sub-sampled (SS)} data, sampled to be as similar as possible to $\mathcal{D}_{L}$'s histogram  
(yielding 70k groups with 250k samples).
Fig. \ref{fig:distributions} provides an illustrations of those histograms.
In order to assign each group of items with one single label, as described earlier
, we choose the category with the highest confidence score within the group provided by $\fbase$ \footnote{Preliminary experiments showed that this method outperformed majority voting.}.


\subsubsection{CGA}

In order to empirically evaluate CGA, we construct $\Dpairs$ from $\mathcal{D}_U$ as described earlier and use a T5 model \cite{raffel2020exploring} (a large Transformer based seq-2-seq model) as $\mathcal{M}$, which we fine-tune on $\Dpairs$ for three epochs.

\textbf{The impact of the filtering score function.} We examine two alternatives of the score function $s$; 1) BLEU score \cite{papineni2002bleu} and 2) a cosine-similarity score that was computed on the output vectors of an all-MinmLM-L6-V2 model \cite{All-MinmLM-L6-V2}. This model maps sentences to a 384 dimensional dense vector space and can be used for tasks such as clustering or semantic search. We compute both scores for each pair of original product title and a corresponding generated title. Preliminary experiments show that filtering by the BLEU score results in a more consistent model. For the rest of the experiments we use the BLEU score as $s$. 
Table \ref{table:generative_examples} contains some examples of generated titles and their corresponding BLEU score.

Using the T5 model, we generate 8 samples based on each sample from $\mathcal{D}_L$, and compute the $s$ 
score of each of those samples. We then perform three experiments to evaluate the impact of the filtering threshold $T$ and the augmentation size $N$. Results are presented in Figure~\ref{fig:t5_exp_graphs}.

\textbf{The impact of the filtering threshold.} For each threshold value $T\in \{0.5,0.6,0.7,0.8\}$, we filter the generated samples. Then, we sub-sample a fixed amount of $N=200k$ samples into $\Daug$ and train a model on $\mathcal{D}_{L}\cup \Daug$.
As $T$ gets higher, the consistency rate of the trained model increases as well, which indicates the need of a filtering phase. 

\textbf{The impact of the augmentation size.} We filter the generated samples using a fixed $T=0.7$. Out of the remaining generated samples, we sub-sample $N\in \{50k,100k,200k,400k\}$ samples into $\Daug$, and train a model on $\mathcal{D}_{L}\cup \Daug$.
As $N$ gets higher, the consistency rate of the trained model increases as well, which indicates that adding more generated samples leads to a more consistent model.

\textbf{The trade off between filtering threshold and augmentation size.}
We filter the generated samples using different thresholds, and add the filtered samples to $\Daug$ without sub-sampling them. We train a model on $\mathcal{D}_{L}\cup \Daug$. 
Evidently, the consistency rate of the trained model increases when $T$ gets higher 
but decreases for $T=0.8$. 
As $T$ gets higher, the filtered samples are of better quality but there are fewer of them, reaching an optimal trade off at $T=0.7$.  
Thus, for the rest of the paper, we use $T=0.7$.


%% file: Chapters/8.results.tex
\section{Results and Discussion}\label{sec:results}

We train each examined model 5 times and present the mean score 
of the achieved results. 
For each model, we compare the weighted average F1 score for the accuracy test and the consistency rate of the consistency test. Table~\ref{tab:exp results} presents our results. 

The ColorsSizes-Blind model performs similarly to the baseline for both measurements; the slight changes are within the std range, thus making the differences insignificant compared to the baseline model. 
This is an evidence that the item versioning is more complex than just changing the size or color and includes title rephrasing concepts that are hard to tackle in a trivial way. 

In addition, the results show that both of the CST versions, complete and sub-sampled, achieve significantly higher consistency rates than the baseline, gaining lifts of 10\% and 7\% respectively. On the other hand, both of the methods yield lower F1 scores, reducing lift by 1.65\% and 0.6\% respectively. 
A possible cause of the degradation in the F1 score is the differences between the data distribution of $\mathcal{D}_L$, which we sample the accuracy test from, and the data distribution of $\mathcal{D}_U$ which we use to augment our training data. \ariel{rephrase} The fact that using the sub-sampled version of $\mathcal{D}_U$ mitigates most of this degradation supports this claim. 
An additional cause could be the usage of the noisy pseudo-labels in the augmented set instead of the unavailable ground truth labels. Note that the amount of added data using $\mathcal{D}_U$ to tackle consistency is bigger than the original $\mathcal{D}_L$, which aims to tackle accuracy. The focus in terms of the training shifts from an accuracy problem to a consistency problem, thus hurting the F1 of the new model. The higher consistency rate of CST-Full compared to the CST-Sub-Sampled can be explained by a difference of more than 1M samples in the size of $\Daug$.  

\tabcolsep=0.11cm
\begin{table}[t]
\centering
\begin{tabular}{|l |c | c |c| c|}
 \hline
Method & F1 & \specialcell{F1 lift} & \specialcell{Cnst.\\rate} & \specialcell{Cnst.\\lift}\\
\hline
Baseline & 0.665 & - & 0.738 & -\\ 
 CS-Blind & $0.664
 $ & -0.13\% & $0.740
 $ & 0.26\%\\ 
 CST-Full & $0.654
 $ & -1.65\% & $\mathbf{0.813
 }$ & \textbf{10.12\%}\\ 
 CST-SS & $0.661
 $ & -0.6\% & $0.790
 $ & 6.99\%\\ 
 CGA & $\mathbf{0.667
 }$ & \textbf{0.28\%} & $0.771
 $ & 4.46\%\\ \hline
\end{tabular}
\caption{Categorization results, indicating the mean. Lift values are all compared to the Baseline model. The std ranges between $0.001$ to $0.002$ for F1 and $0.001$ to $0.009$ for the consistency rate.}
\label{tab:exp results}
\end{table}

Similarly, the CGA method also improves the consistency rate, gaining lift of 4.5\%, and doesn't significantly affect the accuracy score. As mentioned, we use a threshold $T=0.7$, thus including 440k samples in $\Daug$. These additional samples correspond to a similar distribution as $\mathcal{D}_L$.
The improvement in both the consistency and the accuracy indicates that the generative model is able to correctly learn the real-world item versioning and produce a significant amount of data with high accuracy labels and the same distribution as in the accuracy test.

Summarizing the above, our experiments highlight three key factors in the consistent-SSL framework: 1) \textbf{Scale}~- enriching the learning set with more examples of item versioning increases the consistency. 
2) \textbf{Quality}~- augmenting the data with real-world samples is better than using generative ones in term of performance.
3) \textbf{Distribution}~- preserving the original distribution in the augmented set is important for maintaining good accuracy.



%% file: Chapters/9.conclusions.tex
\section{Conclusions}
This work presents a new framework for consistent text categorization in the context of e-Commerce. The aim of this work is to improve a product categorization model that serves various 
services of a major web company. 
We address the labeling inconsistency issues found in the categorization of similar items, leading to poor user experience in related recommendation and search applications. Our framework utilizes an unlabeled clustered dataset in two ways: a self-training approach and a generative-augmentation method.  We performed a thorough investigation of the two approaches and investigated several factors that majorly influence their performance. 
Our experimental results suggest that both proposed methods improve the consistency rate by 4\% to 10\%, while maintaining the accuracy of the current production model.
Finally, our study illustrates 
the trade off between the quality and the scale of the augmented dataset, and its impact on the performance of both methods. 


%% file: Chapters/appendix.tex
\section{CST Algorithm} \label{alg:bootstrapping}

\begin{algorithm}
\caption{CST}

\begin{algorithmic}[1]

\Require{labeled training data $\mathcal{D}_{L}=\{(x_i, y_i)\}_{i=1}^{l}$, unlabeled data $\mathcal{D}_{U}=\{X_i\}_{i=l+1}^{l+u}$, $X_i=\left\{\hat{x}_j^{(i)}\right\}_{j=1}^{k_i}$, set function $h$}

\State \text{train a base classifier $\fbase$ on $\mathcal{D}_{L}$}

\State \text{ $\Daug\gets \emptyset$}

\For {$i=l+1, l+2,\ldots ,l+u$}
    \State {$\tilde{y}_i\gets h(X_i;\fbase)$}
    \State \text{$\Daug \gets \Daug \cup \left\{ (\hat{x}_j^{(i)}, \tilde{y}_i) \right\}_{j=1}^{k_i}$}
\EndFor

\State \text{train $f$ on $\mathcal{D}_{L}\cup \Daug$}

\State \Return $f$
\end{algorithmic}
\end{algorithm}



\section{CGA Algorithm} \label{alg:gen}

\begin{algorithm}
\caption{CGA}

\begin{algorithmic}[1]

\Require {labeled training data $\mathcal{D}_{L}=\{(x_i, y_i)\}_{i=1}^{l}$, unlabeled data $\mathcal{D}_{U}=\{\hat{X}_i\}_{i=l+1}^{l+u}$, $X_i=\left\{\hat{x}_j^{(i)}\right\}_{j=1}^{k_i}$, number of samples to generate from each original sample $n$, score function $s$, threshold $T$ }

\State \text{$\Dpairs = \left\{\left(\hat{x}_j^{(i)}, \hat{x}_{j'}^{(i)}\right)\Big| l+1\leq i\leq l+u\bigwedge j,j'\in [k_i] \right\}$}

\State \text{train a generative model $\mathcal{M}$ on $\Dpairs$}

\State \text{$\Daug\gets \emptyset$}

\For{$i=1,2,\ldots ,l$}
    \State \text{generate $n$ new samples ${\hat{x}}_{1}^{(i)},\ldots,{\hat{x}}_{n}^{(i)}$ with $\mathcal{M}$ and $x_i$}
    
    \For{$j=1,2,\ldots ,n$}
       \If{$s({\hat{x}}_{j}^{(i)}, x_i) \geq T$\\\quad\quad\:\:\:}
            $\Daug \gets \Daug \cup \left\{ ({\hat{x}}_{j}^{(i)}, y_i) \right\}$
        \EndIf
    \EndFor
\EndFor

\State \text{train $f$ on $\mathcal{D}_{L}\cup \Daug$}

\State \Return $f$
\end{algorithmic}
\end{algorithm}

 \section{Hierarchical-FastText}
\label{sec:HFT}
Hierarchical-FastText (HFT) consist of 4 FastText models $\{f_i\}_{i=1}^{4}$.
In training time, each $f_i$ is trained over the same data samples, but with different granularity of the labels: $f_1$ is trained using only the first level of the labels $L_1$, $f_2$ is trained using the first and second levels of the labels $L_1$ and $L_2$ and so on.
In inference time, we use an iterative method, were at each iteration $i$ for $i=1,\ldots ,4$ we predict the label using $f_i$. 
If $f_i$ agrees with $f_{i-1}$ on the label until the level $L_{i-1}$, the process continues, otherwise it returns the prediction of $f_{i-1}$.
If the process gets to the end, i.e. $f_4$ agrees with $f_3$ on the label until $L_3$, it returns the prediction of $f_4$ as the final prediction.

%% file: main.bbl
\begin{thebibliography}{19}
\expandafter\ifx\csname natexlab\endcsname\relax\def\natexlab#1{#1}\fi

\bibitem[{All-MinmLM-L6-V2()}]{All-MinmLM-L6-V2}
All-MinmLM-L6-V2. 2022.
\newblock All-minmlm-l6-v2.
\newblock \url{https://huggingface.co/sentence-transformers/all-MiniLM-L6-v2}.
\newblock [Online; accessed 10-October-2022].

\bibitem[{Arazo et~al.(2020)Arazo, Ortego, Albert, O’Connor, and
  McGuinness}]{arazo2020pseudo}
Eric Arazo, Diego Ortego, Paul Albert, Noel~E O’Connor, and Kevin McGuinness.
  2020.
\newblock Pseudo-labeling and confirmation bias in deep semi-supervised
  learning.
\newblock In \emph{2020 International Joint Conference on Neural Networks
  (IJCNN)}, pages 1--8. IEEE.

\bibitem[{Arjovsky et~al.(2019)Arjovsky, Bottou, Gulrajani, and
  Lopez-Paz}]{arjovsky2019invariant}
Martin Arjovsky, L{\'e}on Bottou, Ishaan Gulrajani, and David Lopez-Paz. 2019.
\newblock Invariant risk minimization.
\newblock \emph{arXiv preprint arXiv:1907.02893}.

\bibitem[{Bari et~al.(2020)Bari, Mohiuddin, and Joty}]{bari2020uxla}
M~Saiful Bari, Tasnim Mohiuddin, and Shafiq Joty. 2020.
\newblock Uxla: A robust unsupervised data augmentation framework for
  zero-resource cross-lingual nlp.
\newblock \emph{arXiv preprint arXiv:2004.13240}.

\bibitem[{Elazar et~al.(2021)Elazar, Kassner, Ravfogel, Ravichander, Hovy,
  Sch{\"u}tze, and Goldberg}]{elazar2021measuring}
Yanai Elazar, Nora Kassner, Shauli Ravfogel, Abhilasha Ravichander, Eduard
  Hovy, Hinrich Sch{\"u}tze, and Yoav Goldberg. 2021.
\newblock Measuring and improving consistency in pretrained language models.
\newblock \emph{Transactions of the Association for Computational Linguistics},
  9:1012--1031.

\bibitem[{Jin et~al.(2020)Jin, Jin, Zhou, and Szolovits}]{jin2020bert}
Di~Jin, Zhijing Jin, Joey~Tianyi Zhou, and Peter Szolovits. 2020.
\newblock Is bert really robust? a strong baseline for natural language attack
  on text classification and entailment.
\newblock \emph{Proceedings of the AAAI conference on artificial intelligence},
  34(05):8018--8025.

\bibitem[{Joulin et~al.(2016)Joulin, Grave, Bojanowski, and
  Mikolov}]{joulin2016bag}
Armand Joulin, Edouard Grave, Piotr Bojanowski, and Tomas Mikolov. 2016.
\newblock Bag of tricks for efficient text classification.
\newblock \emph{arXiv preprint arXiv:1607.01759}.

\bibitem[{Kaushik et~al.(2019)Kaushik, Hovy, and Lipton}]{kaushik2019learning}
Divyansh Kaushik, Eduard Hovy, and Zachary~C Lipton. 2019.
\newblock Learning the difference that makes a difference with
  counterfactually-augmented data.
\newblock \emph{arXiv preprint arXiv:1909.12434}.

\bibitem[{Kumar et~al.(2020)Kumar, Choudhary, and Cho}]{kumar2020data}
Varun Kumar, Ashutosh Choudhary, and Eunah Cho. 2020.
\newblock Data augmentation using pre-trained transformer models.
\newblock \emph{arXiv preprint arXiv:2003.02245}.

\bibitem[{Lee et~al.(2013)}]{lee2013pseudo}
Dong-Hyun Lee et~al. 2013.
\newblock Pseudo-label: The simple and efficient semi-supervised learning
  method for deep neural networks.
\newblock \emph{Workshop on challenges in representation learning, ICML},
  3(2):896.

\bibitem[{Papineni et~al.(2002)Papineni, Roukos, Ward, and
  Zhu}]{papineni2002bleu}
Kishore Papineni, Salim Roukos, Todd Ward, and Wei-Jing Zhu. 2002.
\newblock Bleu: a method for automatic evaluation of machine translation.
\newblock In \emph{Proceedings of the 40th annual meeting of the Association
  for Computational Linguistics}, pages 311--318.

\bibitem[{Raffel et~al.(2020)Raffel, Shazeer, Roberts, Lee, Narang, Matena,
  Zhou, Li, Liu et~al.}]{raffel2020exploring}
Colin Raffel, Noam Shazeer, Adam Roberts, Katherine Lee, Sharan Narang, Michael
  Matena, Yanqi Zhou, Wei Li, Peter~J Liu, et~al. 2020.
\newblock Exploring the limits of transfer learning with a unified text-to-text
  transformer.
\newblock \emph{J. Mach. Learn. Res.}, 21(140):1--67.

\bibitem[{Rizos et~al.(2019)Rizos, Hemker, and Schuller}]{rizos2019augment}
Georgios Rizos, Konstantin Hemker, and Bj{\"o}rn Schuller. 2019.
\newblock Augment to prevent: short-text data augmentation in deep learning for
  hate-speech classification.
\newblock In \emph{Proceedings of the 28th ACM international conference on
  information and knowledge management}, pages 991--1000.

\bibitem[{Shorten et~al.(2021)Shorten, Khoshgoftaar, and
  Furht}]{shorten2021text}
Connor Shorten, Taghi~M Khoshgoftaar, and Borko Furht. 2021.
\newblock Text data augmentation for deep learning.
\newblock \emph{Journal of big Data}, 8(1):1--34.

\bibitem[{Triguero et~al.(2015)Triguero, Garcia, and
  Herrera}]{triguero2015self}
Isaac Triguero, Salvador Garcia, and Francisco Herrera. 2015.
\newblock Self-labeled techniques for semi-supervised learning: taxonomy,
  software and empirical study.
\newblock \emph{Knowledge and Information systems}, 42(2):245--284.

\bibitem[{Veitch et~al.(2021)Veitch, D'Amour, Yadlowsky, and
  Eisenstein}]{veitch2021counterfactual}
Victor Veitch, Alexander D'Amour, Steve Yadlowsky, and Jacob Eisenstein. 2021.
\newblock Counterfactual invariance to spurious correlations: Why and how to
  pass stress tests.
\newblock \emph{arXiv preprint arXiv:2106.00545}.

\bibitem[{Wang et~al.(2020)Wang, Wang, Qin, Packer, Li, Chen, Beutel, and
  Chi}]{wang2020cat}
Tianlu Wang, Xuezhi Wang, Yao Qin, Ben Packer, Kang Li, Jilin Chen, Alex
  Beutel, and Ed~Chi. 2020.
\newblock Cat-gen: Improving robustness in nlp models via controlled
  adversarial text generation.
\newblock \emph{arXiv preprint arXiv:2010.02338}.

\bibitem[{Wang et~al.(2021)Wang, Yang, and Wang}]{wang2021identifying}
Tianlu Wang, Diyi Yang, and Xuezhi Wang. 2021.
\newblock Identifying and mitigating spurious correlations for improving
  robustness in nlp models.
\newblock \emph{arXiv preprint arXiv:2110.07736}.

\bibitem[{Xie et~al.(2020)Xie, Dai, Hovy, Luong, and Le}]{xie2020unsupervised}
Qizhe Xie, Zihang Dai, Eduard Hovy, Thang Luong, and Quoc Le. 2020.
\newblock Unsupervised data augmentation for consistency training.
\newblock \emph{Advances in Neural Information Processing Systems},
  33:6256--6268.

\end{thebibliography}
